%% file: ipm.tex
\documentclass[review]{elsarticle}

\usepackage{lineno,hyperref}
\usepackage{bm}
\usepackage{amsmath}
\usepackage{amssymb}
\usepackage{bbm}

\modulolinenumbers[5]

\begin{document}

\begin{frontmatter}

\title{Read Beyond the Lines: Understanding the Implied Textual Meaning via A Skim and Intensive Reading Model}

\author{Guoxiu He\footnote[1]{Work done as an intern at Alibaba Group.}}
\address{Wuhan University, Wuhan, China, guoxiu.he@whu.edu.cn}

\author{Zhe Gao}
\address{Alibaba Group, Hangzhou, China, gaozhe.gz@alibaba-inc.com}

\author{Zhuoren Jiang}
\address{Sun Yat-sen University, Guangzhou, China, jiangzhr3@mail.sysu.edu.cn}

\author{Yangyang Kang}
\address{Alibaba Group, Hangzhou, China, yangyang.kangyy@alibaba-inc.com}

\author{Changlong Sun}
\address{Alibaba Group, Hangzhou, China, changlong.scl@taobao.com}

\author{Xiaozhong Liu\footnote[2]{Corresponding authors.}}
\address{Indiana University Bloomington, Bloomington, United States, liu237@indiana.edu}

\author{Wei Lu\footnotemark[2]}
\address{Wuhan University, Wuhan, China, weilu@whu.edu.cn}







\begin{abstract}
The nonliteral interpretation of a text is hard to be understood by machine models due to its high context-sensitivity and heavy usage of figurative language. In this study, inspired by human reading comprehension, we propose a novel, simple, and effective deep neural framework, called \textbf{S}kim and \textbf{I}ntensive \textbf{R}eading \textbf{M}odel (SIRM), for figuring out implied textual meaning. The proposed SIRM consists of two main components, namely the skim reading component and intensive reading component. N-gram features are quickly extracted from the skim reading component, which is a combination of several convolutional neural networks, as skim (entire) information. An intensive reading component enables a hierarchical investigation for both local (sentence) and global (paragraph) representation, which encapsulates the current embedding and the contextual information with a dense connection. More specifically, the contextual information includes the near-neighbor information and the skim information mentioned above. Finally, besides the normal training loss function, we employ an adversarial loss function as a penalty over the skim reading component to eliminate noisy information arisen from special figurative words in the training data. To verify the effectiveness, robustness, and efficiency of the proposed architecture, we conduct extensive comparative experiments on several sarcasm benchmarks and an industrial spam dataset with metaphors. Experimental results indicate that (1) the proposed model, which benefits from context modeling and consideration of figurative language, outperforms existing state-of-the-art solutions, with comparable parameter scale and training speed; (2) the SIRM yields superior robustness in terms of parameter size sensitivity; (3) compared with ablation and addition variants of the SIRM, the final framework is efficient enough.

\end{abstract}


\end{frontmatter}


\input{ipm-01-intro}
\input{ipm-05-relate}
\input{ipm-02-model}

\input{ipm-03-exp}

\input{ipm-04-res}

\input{ipm-06-conclu}
\input{ipm-08-acknowledge}


\bibliography{ipm}

\end{document}

%% file: ipm-01-intro.tex
\section{Introduction}
\label{sec:intro}

\begin{figure}[!ht]
\centering
\includegraphics[width=8cm]{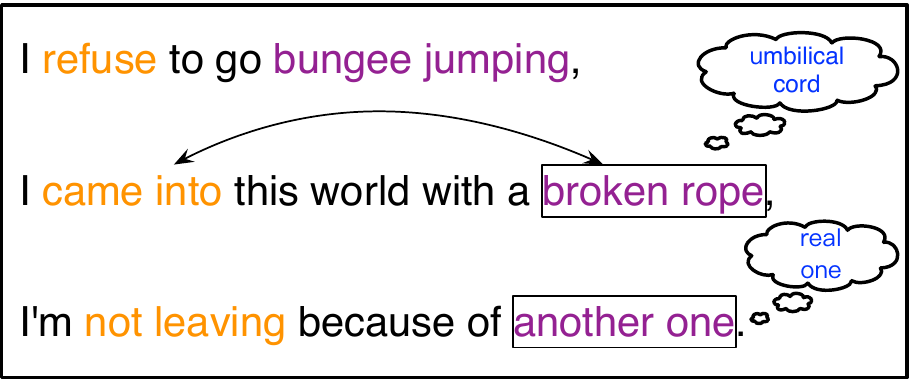}
\caption{This is a typical case, which is a sentence expresses non-literal meaning.}
\label{fig:case}
\end{figure}

Language does not always express its literal meaning, e.g., sarcasm and metaphor. People often use words that deviate from their conventionally accepted definitions in order to convey complicated and implied meanings \cite{tay2018reasoning}. A typical example is shown in Figure \ref{fig:case}.

Compared with standard (literal) text usage, the non-literal text can be associated with two typical linguistic phenomena: 

\begin{itemize}
  \item From syntax and semantic viewpoint, the non-literal text is \textbf{highly context-sensitive}. People can perceive the implied meaning of text through unnatural language usage in context. For instance, as Figure \ref{fig:case} shows, \textit{`rope'} has two different meanings: literal meaning for \textit{`bungee jumping'} and implied meaning (\textit{`umbilical cord'}) for \textit{`came into this world'}. People can tell the difference after digesting the whole sentence.
  \item From a lexicon viewpoint, the non-literal text is often created by presenting words that are equated, compared, or associated with \textbf{normally unrelated or figurative meanings}. These words express different or even opposite meanings, which could change the word distribution under a semantic topic or sentiment polarity and then hinder the training of machine models. In addition, some of these words frequently appearing in the training set will mislead the machine model in the inference process. For instance, no matter how many times \textit{`rope'} refers to \textit{`umbilical cord'} in the training set, we can not assert that it conveys the same meaning in a new coming text.
\end{itemize}

Existing text representation studies, which mainly rely on content embeddings \cite{mikolov2013distributed} based deep neural networks \cite{lecun2015deep} such as Recurrent Neural Networks (RNNs), Convolutional Neural Networks (CNNs), and Attention Mechanisms, are not totally suitable for aforementioned problems. RNNs, such as Long Short Term Memory (LSTM) \cite{hochreiter1997long}, Gated Recurrent Unit (GRU) \cite{cho2014learning}, and Simple Recurrent Unit (SRU) \cite{lei2018sru} draw on the idea of the language model \cite{bengio2003neural}. However, RNNs, including bidirectional ones, could neglect the long-term dependency, as demonstrated in \cite{wang2016inner,pmlr-v80-li18c,shen2018ordered}, since the current term directly depends on the previous term as opposed to the entire information. Although attention mechanisms \cite{yang2016hierarchical} over RNNs provide an important potential to aggregate all hidden states, they focus more on the local part of a text. CNNs \cite{kalchbrenner2014convolutional} can characterize local spatial features and then assemble these features by deeper layers which are expert in extracting phrase-level features. Self-attention mechanism \cite{vaswani2017attention} characterizes the dependency on one term with others in the input sequence and then encodes the mutual relationship to capture the contextual information. Unfortunately, all standard text representation models have not effectively utilized the contextual representation as input directly when encoding the current term, which is necessary to understand the implied meaning. There are also several tailored models for sarcasm detection \cite{tay2018reasoning}, which concentrate more on word incongruity in the text. 


\subsection{Research Objectives}

Hence, the study aims to cope with the following drawbacks which can be summarized as follows:

\begin{itemize}
  \item Existing text representation models don't specifically design a mechanism to effectively use context/global information when understanding the implied meaning of the input text.
  \item Meanwhile, all existing models neglect the potential bad effect of the figurative words which can be frequently appearing in the training set of implied texts.
  \item Existing methods do not take into account both model complexity and model performance at the time of design, which can be very important in practical applications.
\end{itemize}


To this end, we try to design a simple and effective model to interpret the implied meaning by overcoming the challenges mentioned above. From a human reading comprehension viewpoint, to understand a difficult text, a human may firstly skim it quickly to estimate the entire information of the target text. Then, in order to consume the content, he/she can read the text word by word and sentence by sentence with respect to the entire information. Inspired by the above procedure, in this study, we propose a novel deep neural network, namely Skim and Intensive Reading Model (SIRM), to address the implied textual meaning identifying problem. Furthermore, a human can skip some noisy or unimportant figurative phrases seen before to quickly consume the main idea of the target text. An adversarial loss is used in SIRM to achieve the same effect. To the best of our knowledge, SIRM is the first model trying to simulate such human reading procedure for understanding and identifying the non-literal text.

Furthermore, taking efficiency into consideration, we design the details of the proposed SIRM under the Occam's Razor: \textit{`More things should not be used than are necessary'}. In other words, under the premise of optimal task performance, we will remove unnecessary components and use the simplest architecture.

\subsection{Contributions}

Briefly, given the above research objectives, our main contributions of this work can be summarized as follows:
\begin{itemize}
    \item The challenges of understanding the implied textual meaning is well investigated in this research, which can be summarized as follows: context-sensitive and usage of figurative meaning. To the best of our knowledge, these challenges have not been thoroughly studied.
    \item We propose the SIRM to understand implied textual meaning where the intensive reading component, which enables a hierarchical investigation for sentence and paragraph representation, depends on the global information extracted by the skim reading component. The cooperation of the skim reading component and the intensive reading component in the SIRM achieves a positive impact on comprehending nonliteral interpretation by modeling the global contextual information directly. 
    \item We introduce an adversarial loss as a penalty over the skim reading component to cut down noise due to special figurative words during the training procedure.
    \item We conduct extensive comparative experiments to show the effectiveness, robustness, and efficiency of the SIRM. Compared with the existing alternative models, the SIRM achieves superior performance on F1 score and accuracy with a comparable parameter size and training speed. In addition, the SIRM outperforms all other models according to model robustness. And the ablation and addition tests show that the final SIRM is efficient enough.
\end{itemize}

The remainder of this paper is structured as follows: Some related works are summarized in Section \ref{sec:related} and the details of the proposed SIRM are introduced in Section \ref{sec:model}. Section \ref{sec:eval} presents the experimental settings which is followed by results and analyses in Section \ref{sec:result}. Our concluding remarks in Section \ref{sec:conclu}.

%% file: ipm-05-relate.tex
\section{Related Work}
\label{sec:related}

This work is related to deep neural networks and semantic representation for text understanding.

Recently, a large number of CNNs and RNNs with potential benefits have attracted many researchers' attention. Existing efforts mainly focus on the application of LSTM~\cite{hochreiter1997long,pichotta2016using,palangi2016deep}, GRU~\cite{cho2014learning,chung2014empirical}, SRU~\cite{lei2018sru}, and CNNs~\cite{kalchbrenner2014convolutional,gehring2017convs2s,johnson2017deep} based on word embeddings~\cite{mikolov2013distributed,Mikolov2013Efficient} drawing on the idea of either language model~\cite{bengio2003neural,mikolov2010recurrent} or spatial parameter sharing. And all these models have demonstrated impressive results in NLP applications. Many previous works have shown that the performance of deep neural networks can be improved by attention mechanism~\cite{bahdanau2014neural,he2018entire}. In addition, self-attention mechanism with position embedding characterizes the mutual relationship between one and others as dependency to capture the semantic encoding information~\cite{vaswani2017attention}. There are some other works that combine RNN and CNN for text classification~\cite{zhou2015c,wang2017hybrid} or use hierarchical structure for language modeling~\cite{lin2015hierarchical,yang2016hierarchical}. Besides hybrid neural networks, graph based models~\cite{noekhah2020opinion,jiang2019detect} and human behavior enhanced models~\cite{he2019finding} are widely employed to capture textural semantics.

Recently, sarcasm detection, one of the implied semantic recognitions, is widely studied by linguistic researchers~\cite{kunneman2015signaling,camp2012sarcasm,campbell2012there,ghosh2016fracking,ivanko2003context}. \cite{fersini2016expressive} proves that it is important to consider several valuable expressive forms to capture the sentiment orientation of the messages. And external sentiment analysis resources are benefit for sarcasm detection \cite{zhang2019irony}. Furthermore, \cite{tay2018reasoning} realizes a neural network to represent a sentence by comparing word-to-word embeddings which achieves the state-of-the-art performance. More specifically, an intra-attention mechanism allows their model to search for conflict sentiments as well as maintain compositional information. 

However, all these approaches mentioned above don't specifically make good use of the contextual representation as a direct input when understanding the implied meaning and they never worry about the possible noise such as special figurative phrases in the training data.

%% file: ipm-02-model.tex
\section{Skim and Intensive Reading Model}
\label{sec:model}

\begin{figure}[!ht]
\centering
\includegraphics[width=9cm]{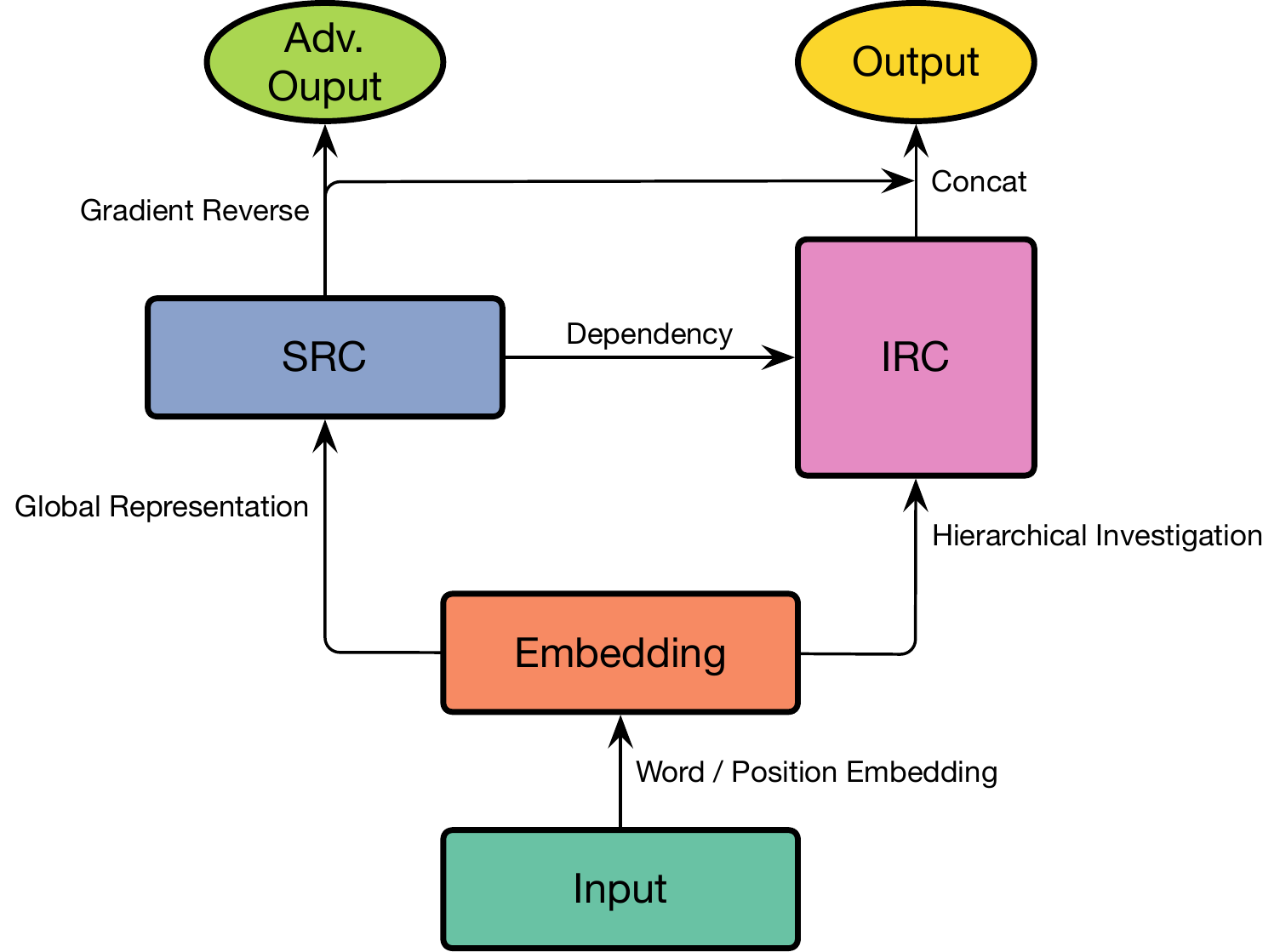} 
\caption{This is the architecture of the proposed SIRM, mainly including a skim reading component, an intensive reading component, and an adversarial loss. Each part of the SIRM is designed under the Occam's razor.}
\label{fig:sirm}
\end{figure}

In this section, we propose a novel deep neural network inspired by reading comprehension procedure of people, namely Skim and Intensive Reading Model~(SIRM), to address the essential issues for understanding texts with implied meanings. The architecture of the model is depicted in Figure~\ref{fig:sirm}. 

\subsection{Overview}
People always consume a difficult text word by word and sentence by sentence with respect to the global information extracted by reading quickly. Besides the input layer and the embedding layer, the SIRM consists of two main parts which are the skim reading component~(SRC) associated with an adversarial loss and the intensive reading component~(IRC) simulate the procedure of human reading comprehension. And for efficiency concern, the model is designed under the Occam's razor, which means we use the simplest and minimal component to realize each part of the SIRM. More specifically, the SRC is a set of shallow CNNs to enable global feature extraction, while the IRC is a hierarchical framework to enhance the contextual information, from sentence level to paragraph level. Finally, over the output layer, a common cross entropy and an adversarial loss are utilized to represent the cost function of the end-to-end deep neural network.

\subsection{Input}
Each example of this task is represented as a set $(p, y)$, where input $p=[s_1, \cdots, s_m]$ is a paragraph with $m$ sentences, $s_i=[w_{i,1}, \cdots, w_{i,n}]$ is $i-$th sentence in paragraph $p$ with $n$ words, and $y\in Y$ where $Y=\{0, 1\}$ is the label representing the category of $p$. We can represent the task as estimating the conditional probability $Pr(y|p)$ based on the training set, and identifying whether  a testing example belongs to the target class by $y' = argmax_{y\in Y} Pr(y|p)$.

\subsection{Word Embedding}
The goal of word embedding layer is to represent $j-$th word $w_{i,j}$ in sentence $s_i$ with a $d_e$ dimensional dense vector $\bm{x}_{ij} \in \mathbb{R}^{d_e}$. Given an input paragraph $p$, it will be represented as $\bm{P}=[\bm{S}_1, \cdots, \bm{S}_m]\in \mathbb{R}^{m\times n\times d_e}$, where each representation of sentence is a matrix $\bm{S}_i=[\bm{x}_{i,1}, \cdots, \bm{x}_{i,n}]\in \mathbb{R}^{n\times d_e}$ consisting of word embedding vectors of $i-$th sentence.

\subsection{Position Embedding}
Position information can be potentially important for text understanding. In the SIRM, two types of position information are encoded, word position in a sentence, and sentence position in a paragraph. By leveraging the position encoding method from~\cite{vaswani2017attention}, word/sentence positions are captured via sine and cosine functions of different frequencies to the input embeddings. Furthermore, the positional encodings have the same dimension as the corresponding embedding matrix, so that the results can be easily aggregated. The mathematical formulas are shown as follows:
\begin{equation}
\begin{split}
	& \bm{U}_{pos, 2i}=sin(pos/10000^{2i/d_e})~, \\
	& \bm{U}_{pos, 2i+1}=cos(pos/10000^{2i/d_e})~,
\end{split}
\end{equation}
where $pos$ is the position and $i$ denotes the dimension. Moreover, for any fixed offset $k$, $\bm{U}_{pos+k}$ can be represented as a sinusoidal function of $\bm{U}_{pos}$.

After that, we add corresponding position embedding matrix $\bm{U}_{1:n}$ to each sentence embedding matrix $\bm{S}_i$:
\begin{equation}
\begin{split}
	& \bm{S}'_i = \bm{S}_i + \bm{U}_{1:n}=[\bm{x}_{i,1}', \bm{x}_{i,2}', \cdots, \bm{x}_{i,n}']~.
\end{split}
\end{equation}

\subsection{Skim Reading Component~(SRC)}

\begin{figure}[!ht]
\centering
\includegraphics[width=12.5cm]{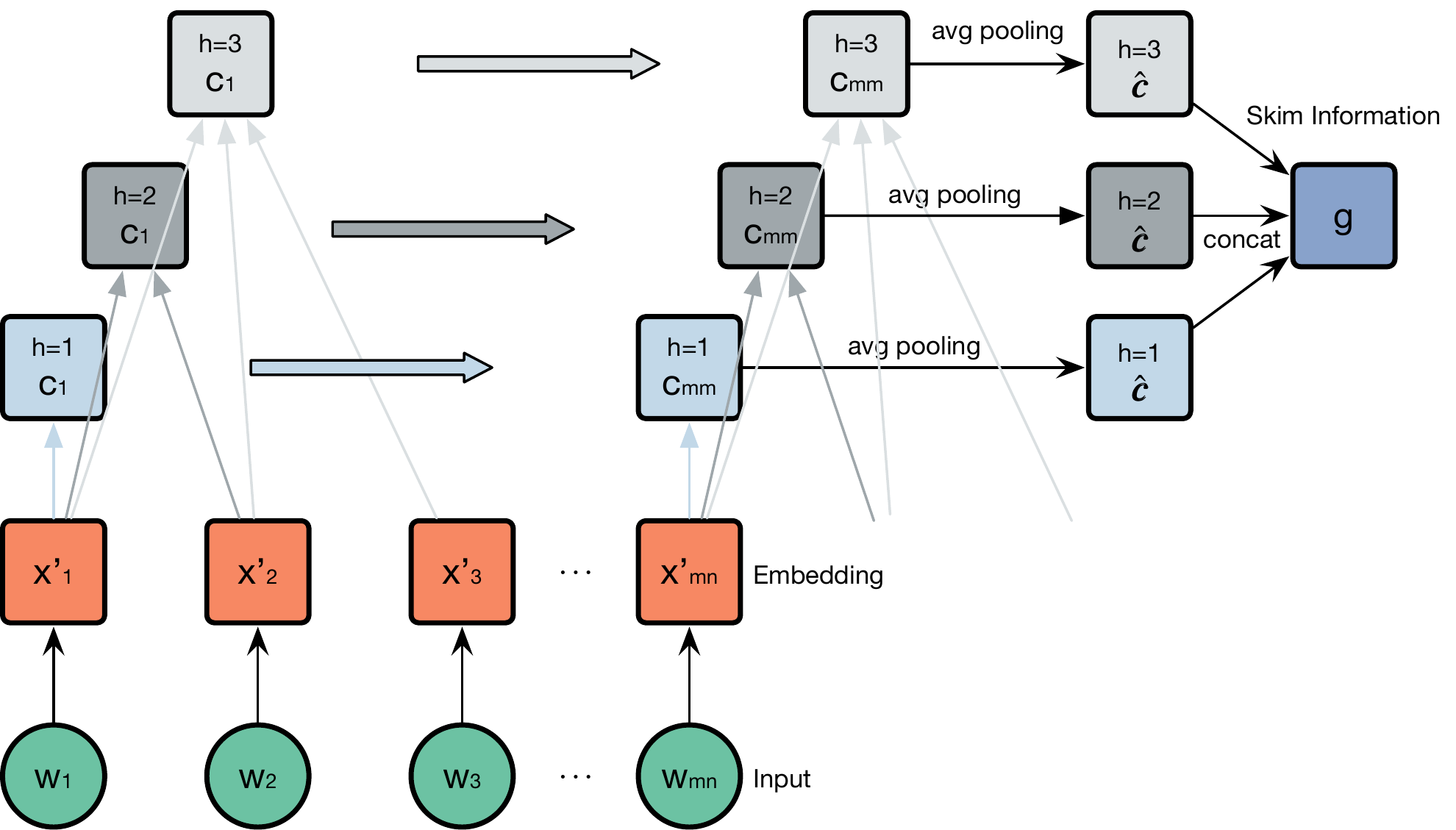} \caption{The SRC characterizes the entire information via convolutional neural networks with different kernel/window size.}
\label{fig:SRC}
\end{figure}

Since each word and sentence with implied textual meaning can be highly dependent on the contextual information, the proposed model needs to characterize the dynamic entire information with a quick manner like a human shown in Figure~\ref{fig:SRC}.

A tailored CNN employs three key functions, e.g., sparse interaction, parameter sharing, and equivariant representation~\cite{wang2017hybrid}, which can encode the partial spatial information. Hence, in the SRC, we use CNN layers with different window sizes in order to extract features like n-gram. Given a paragraph embedding $\hat{\bm{P}}\in \mathbb{R}^{m\cdot n\times d_e}$ reshaped from $\bm{P}$, the global feature is extracted as follows:
\begin{equation}
	\bm{g}=SRC(\hat{\bm{P}})~.
\end{equation}

More specifically, $d_c$ convolution filters are applied to a window of $h$ words to produce a corresponding local feature. For example, a feature $\bm{c}_i\in \mathbb{R}^{d_c}$ is generated from a window of words $\hat{\bm{P}}_{i:i+h-1}$:
\begin{equation}
	\bm{c}_i=ReLU(\bm{W}_{c} * \hat{\bm{P}}_{i:i+h-1} + \bm{b}_{c})~,
\end{equation}
where $*$ denotes the convolutional operation and the feature map from filter with the same shape is represented as $\bm{C}=[\bm{c}_1, \bm{c}_2, \cdots, \bm{c}_{m\cdot n-h+1}]$. 

We then apply an average-over-time pooling operation over the feature map and obtain the feature as follows:
\begin{equation}
	\hat{\bm{c}}=\frac{1}{m\cdot n-h+1}\sum_{i}^{m\cdot n-h+1}\bm{c}_i~.
\end{equation}

In this part, we utilize filters with $c$ kinds of window size to extract more accurate relevant information by taking the consecutive words~(e.g., n-gram) into account, and then concatenate all $\hat{\bm{c}}$ from these filters to get the global semantic feature mentioned above which is represented as $\bm{g}$, where $\bm{g}\in \mathbb{R}^{c\cdot d_c}$.

\subsection{Intensive Reading Component~(IRC)}

\begin{figure}[!ht]
\centering
\includegraphics[width=7.5cm]{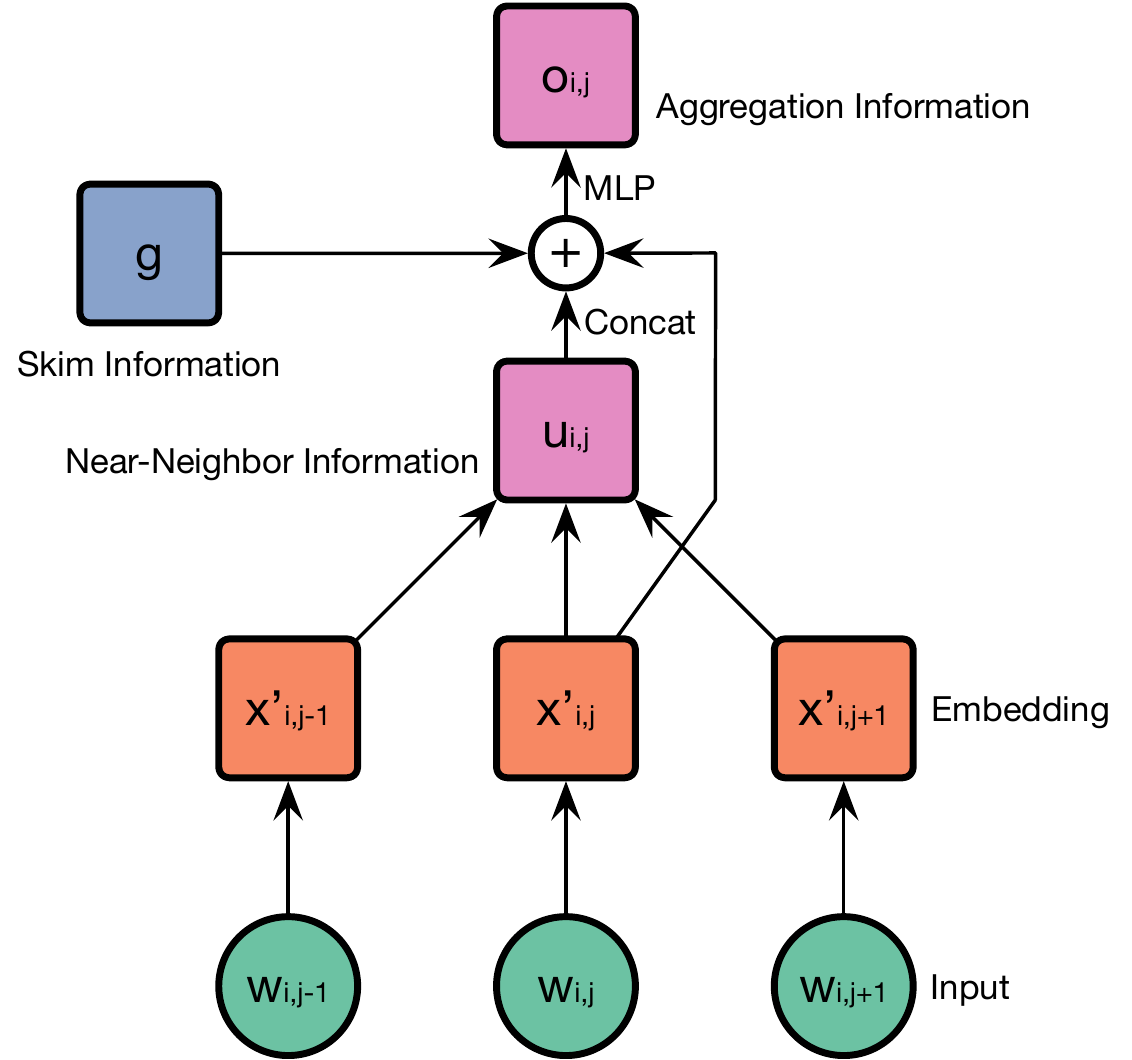} 
\caption{The IRC encodes the current embedding, the near-neighbor information, and the skim information with a dense connection.}
\label{fig:IRC}
\end{figure}

Inspired by the human reading comprehension procedure, the IRC employs a hierarchical framework to characterize and explore the implied semantic information from sentence level to paragraph level. In other words, the sentence encoding outcomes will be used as the input of the paragraph-level part. The structure of IRC is shown in Figure~\ref{fig:IRC}.

For sentence-level part~($IRC_S$), given $i-$th sentence embedding $\bm{S'}_i$ from embedding layer with position embedding and the global information $\bm{g}$ from the SRC, the sentence encoding information is extracted as a vector shown below:
\begin{equation}
	\bm{o}_i=IRC_S(\bm{S'}_i, \bm{g})~,
\end{equation}
and the paragraph embedding information is represented as a matrix: $\bm{O}=[\bm{o}_1, \bm{o}_2, \cdots, \bm{o}_m]$.

Before paragraph-level model~($IRC_P$), a corresponding position embedding matrix $\bm{U}$ is added to $\bm{O}$:
\begin{equation}
	\bm{O}'=\bm{O}+\bm{U}_{1:m}~.
\end{equation}

Then, the paragraph is encoded as a vector shown below:
\begin{equation}
	\bm{o}_P=IRC_P(\bm{O}', \bm{g})~.
\end{equation}

Note that both $IRC_S$ and $IRC_P$ share the same structure, but the trainable parameter values are quite different. The detailed component descriptions can be found as follows.

\subsubsection{Near-Neighbor Information Encoder}
For people, in order to understand the implied meaning of the current word/sentence, besides the entire information of the whole paragraph, the near-neighbor information around the word/sentence, in a size $2\cdot k+1$ word/sentence window, also plays an important role in characterizing the contextual information of the target word/sentence. 

Hence, we pad both $k$ words/sentences at the head and tail for input sentence embedding $\bm{S'}_i$ or paragraph embedding $\bm{O'}$, respectively. Taking the sentence-level part as an example, $d_{ns}$ filters, with window size $2\cdot k + 1$, are applied to produce the near-neighbor information. So, the near-neighbor information of $j-$th word in $i-$th sentence is represented as a vector $\bm{u}_{ij}\in \mathbb{R}^{d_{ns}}$:
\begin{equation}
	\bm{u}_{i,j}=f(\bm{W}_{ns} * \bm{S'}_{i,j-k:j+k}+\bm{b}_{ns})~.
\end{equation}
%

Finally, the near-neighbor information of all words in $i-$th sentence is encoded as a matrix: $\bm{U}_i=[\bm{u}_{i,1}, \bm{u}_{i,2}, \cdots, \bm{u}_{i,n}]$. The near-neighbor information is an important part of contextual information for the current word.

\subsubsection{Dense Connection}

To comprehensively understand implied semantics of a given text, the main effort of this work is to take advantage of the contextual information as a guidance and dependency to each word/sentence like people always do. Hence, inspired by \cite{huang2017densely}, the most direct idea is to concatenate the entire information (the skim information), the near-neighbor information, and the pure word/sentence embedding, and then feed them into a Multilayer Perceptron~(MLP), also named dense connection layer, to realize an aggregate encoding. Taking the sentence-level part as an example, the aggregate encoding is achieved as below:

\begin{equation}
\begin{split}
	& \bm{t}_{i,j} = [\bm{g}\oplus \bm{u}_{i,j} \oplus \bm{x}'_{i,j}]~,\\
	& \bm{o}_{i,j} = relu(\bm{W}_{ls}\cdot \bm{t}_{i,j}+\bm{b}_{ls})~,
\end{split}
\end{equation}
where $\oplus$ is the concatenation operation. 

Eventually, $i-$th sentence from sentence-level model is encoded as $\bm{o}_i\in \mathbb{R}^{d_{as}}$:
\begin{equation}
	\bm{o}_i = \frac{1}{n} \sum_j^{n} \bm{o}_{i,j}~.
\end{equation}

For paragraph-level IRC, the outputs from the near-neighbor information encoder and the aggregate encoder are represented as $\bm{U}_P \in \mathbb{R}^{m\times d_{np}}$ and $\bm{o}_P \in \mathbb{R}^{d_{ap}}$, respectively.

Note that, a gate mechanism~\cite{cho2014learning} could replace the dense connection and an attention mechanism~\cite{yang2016hierarchical} could replace the last average pooling. The results of comparison are shown in Figure~\ref{fig:sub_and_add}.

\subsection{Output}
Undertaking the paragraph encoding $\bm{g}$ and $\bm{o}_P$ from the SRC and IRC respectively, a Multilayer Perceptron~(MLP) is applied to generate the output $y'$:
\begin{equation}
	y' = \sigma(\bm{W}_{o}\cdot (\bm{o}_P\oplus \bm{g}) + \bm{b}_{o})~.
\end{equation}
%

Here, the output $y'$ is the probability of the target category.

\subsection{Model Training with Adversarial Learning}
In the SIRM, the skim information $\bm{g}$ is extracted from a set of shallow CNNs. Because this feature is similar to n-gram instead of deep semantic representation, it can be polluted by the noisy information such like special phrases highly related to the training data, e.g., some special figurative phrases. 

Hence, the proposed model should be able to penalize the features strongly associated with the training data, while the general features should be boosted for the IRC optimization.

In this study, we implement this idea by utilizing an adversarial learning mechanism when training the model. For more theoretical details, refer to~\cite{goodfellow2014generative,ganin2016domain,pmlr-v37-ganin15,miyato2016adversarial}. Specifically, we add a MLP over the SRC shown as follows:
\begin{equation}
	y'' = softmax(\bm{W}_{g}\cdot \bm{g} + \bm{b}_{g})~.
\end{equation}
%

Since the n-gram based global feature tends to be overfitting during training procedure, we expect $\bm{g}$ to have a bit low performance when directly connecting to the output.

In a word, the final loss needs to minimize the normal loss and maximize the adversarial learning based loss, which is represented as:
\begin{equation}
	\zeta(y, y', y'') = min \zeta(y, y') + max \lambda\cdot\zeta_{adv}(y, y'')~,
\end{equation}
where both $\zeta$ and $\zeta_{adv}$ are the negative log likelihood and $\lambda$ is an adjustment factor which is far less than 1. In addition, $\zeta_{adv}$ is named as adversarial loss~(Adv) in this paper.

The SIRM is an end-to-end deep neural network, which can be trained by using stochastic gradient descent~(SGD) methods, such as Adam~\cite{kingma2014adam}. More implementation details will be given in the experiments section. In addition, all $\bm{W}$ and $\bm{b}$ mentioned above are weight matrix and bias respectively.

%% file: ipm-03-exp.tex
\section{Experiments}
\label{sec:eval}

In this section, we conduct extensive experiments to evaluate the proposed SIRM against baseline models and several variants of SIRM. As a byproduct of this study, we release the codes and the hyper-parameter settings to benefit other researchers.

\subsection{Datasets}

In order to validate the performance of the proposed SIRM and make it comparable with alternative baseline models, we conduct our experiments on three publicly available benchmark datasets about sarcasm detection and one real-world industrial spam detection dataset with metaphor. Details for all datasets are summarized in Table~\ref{tab:dataset} and described as below:

\begin{table}[!ht]
\centering
\caption{Statistics for all datasets: $l$ is the length of text and +/- is the proportion of positive and negative samples.}
\label{tab:dataset}
    \begin{tabular}{l|ccc|ccc|c}
        \hline
        Name & Train Size & Test Size & Total Size & Max $l$ & Min $l$ & Avg $l$ & +/- \\
        \hline
        Tweets/ghosh & 50,736 & 3,680 & 54,416 & 56 & 6 & 17 & 1/1 \\
        Reddit/movies & 13,535 & 1,504 & 15,039 & 129 & 6 & 13 & 1/1 \\
        IAC/v1 & 1,483 & 371 & 1,854 & 1,045 & 6 & 57 & 1/1 \\
        Industry/spam & 20,609 & 6,871 & 27,480 & 3,447 & 149 & 393 & 1/3 \\
        \hline
    \end{tabular}
\end{table}

\begin{itemize}
  \item Sarcasm Benchmark Datasets: Followed by previous works, we use Tweets/ghosh\footnote{\url{https://github.com/AniSkywalker/SarcasmDetection}} collected by~\cite{ghosh2016fracking,ghosh2017magnets} from Tweets, Reddit/movies\footnote{\url{http://nlp.cs.princeton.edu/SARC/0.0/}} collected by~\cite{khodak2018large} from Reddit, and IAC/v1\footnote{\url{https://nlds.soe.ucsc.edu/sarcasm1}} collected from Internet Argument Corpus~(IAC) by~\cite{walker2012corpus} in this study.
  \item Industrial Dataset: We also evaluate the performance of the proposed SIRM on a Chinese online novel collection about spam detection with metaphor. The spam novels are firstly complained/reported by readers, e.g., parents of children/teenagers, and then confirmed by auditors. Note that the authors of these novels may purposely avoid using the explicit and sensitive words instead of figurative words because of the censorship.
\end{itemize}


\subsection{Baselines}
We employ the following baseline models~(also see Table~\ref{tab:results2}) for comparison, including word embedding~\cite{mikolov2013distributed} based shallow neural networks, deep learning based models, and recent state-of-the-art models:

\textbf{NBOW}~\cite{shen2018baseline}: is a simple model based on word embeddings with average pooling.

\textbf{CNN}~\cite{kim2014convolutional}: is a simple CNN model with average pooling using different kernels. There are 7 kinds of filters whose widths are from 1 to 7 and each has 100 different ones.

\textbf{LSTM}~\cite{hochreiter1997long}: is a vanilla Long Short-Term Memory Network. We set the LSTM dimension to 100. 

\textbf{Atten-LSTM}~\cite{yang2016hierarchical}: is a LSTM applying an attention mechanism. The dimension is set to 100.

\textbf{GRNN}~\cite{zhang2016tweet}: employs a gated pooling method and a standard pooling method to extract content features and contextual features respectively from a gated recurrent neural network. This model has been demonstrated improvement compared to feature engineering based traditional models for sarcasm detection.

\textbf{SIARN} and \textbf{MIARN}~\cite{tay2018reasoning}: capture incongruities between words with an intra-attention mechanism. A single-dimension intra-attention and a multi-dimension one are employed by SIARN and MIARN respectively. Both of them are the state-of-the-art models for sarcasm detection. We use the default settings by the authors.

\textbf{Self-Atten}~\cite{vaswani2017attention}: is the state-of-the-art model from Google to encode deep semantic information using self-attention mechanism\footnote{\url{https://github.com/tensorflow/models/tree/master/official}}. For the feasibility of training because of the large scale parameters, we set all dimensions as 64 just like ours and other hyper-parameters are same as given settings. 


\subsection{Evaluation Metrics}
We choose to report parameter size~(Param) and running time~(Time) for evaluating the efficiency of the proposed SIRM. More specifically, the unit of the parameter size is thousand, and the whole running time of the NBOW is selected as the unit of Time. For effectiveness, we select Macro-Averaged F1 score~(M F1) to show the performance for the label-balanced datasets and employ F1 score~(F1) for label-unbalanced dataset. In addition, we report accuracy~(Acc) for all of datasets. 

\subsection{Experiment Settings}
For experiment fairness, we exploit the same data preprocessing as~\cite{tay2018reasoning}. For the SIRM, the number of convolution filters $d_c$ in the SRC is 16 and the window size $h$ is from 1 to 4. The near-neighbor size $k$ is 1. The dimension $d$ of all other layers are all set to 64. The adjustment factor $\lambda$ for adversarial loss is $1\times 10^{-6}$. In addition, the learning rate is $1\times 10^{-3}$ and the batch size is 64. For Chinese novel dataset, we use \textit{JIEBA}\footnote{\url{https://github.com/fxsjy/jieba}} for tokenization. Furthermore, the statistical significance is conducted via the t-test with p-value$<10^{-3}$.

%% file: ipm-04-res.tex
\section{Results and Analysis}
\label{sec:result}

In this section, we give detailed experimental results and analysis to show insights into our model.

%
%

\subsection{Performance Comparison}


The parameter size, the running time, and the performance of the SIRM compared with baseline models are shown in Table~\ref{tab:results1} and Table~\ref{tab:results2}.

\begin{table}[!ht]
\centering
\caption{Experimental results of efficiency comparison.}
\label{tab:results1}
    \begin{tabular}{l|c|c}
        \hline
        Model & Param & Time\\
        \hline
        NBOW & \textbf{10.3} & \textbf{1} \\
        \hline
        CNN & \underline{30.3} & \underline{2} \\
        LSTM & 60.6 & 18 \\
        Atten-LSTM & 71.0 & 22 \\
        GRNN & 131.0 & 33 \\
        \hline
        SIARN & 100.9 & 150 \\
        MIARN & 102.3 & 180 \\
        Self-Atten & 254.9 & 17 \\
        \hline
        SIRM & 63.7 & \underline{2} \\
        \hline
    \end{tabular}
\end{table}

\begin{table}[!ht]
\centering
\caption{Experimental results of performance comparison.}
\label{tab:results2}
    \begin{tabular}{l|cc|cc|cc|cc}
        \hline
        & \multicolumn{2}{|c|}{Tweets/gosh} & \multicolumn{2}{|c|}{Reddit/movies} & \multicolumn{2}{|c|}{IAC/v1} & \multicolumn{2}{|c}{Industry/spam} \\
        \hline
        Model & M F1 & Acc & M F1 & Acc & M F1 & Acc & F1 & Acc \\
        \hline
        NBOW & 72.42 & 69.37 & \underline{68.50} & \underline{68.18} & 61.32 & 59.61 & 84.96 & 92.25 \\
        \hline
        CNN & 74.84 & 74.54 & 65.50 & 65.03 & 60.98 & 58.40 & 85.46 & 92.42 \\
        LSTM & 75.08 & 75.16 & 67.71 & 66.74 & 44.73 & 53.84 & 82.16 & 90.86 \\
        Atten-LSTM & 75.15 & 73.73 & 65.20 & 63.84 & \underline{61.80} & 60.46 & 83.74 & 91.40 \\
        GRNN & 79.43 & 79.24 & 64.59 & 63.19 & 52.45 & 54.78 & 86.30 & 93.04 \\
        \hline
        SIARN & \underline{78.84} & \underline{79.59} & 67.50 & 68.17 & 60.86 & \underline{61.33} & 77.73 & 92.91\\
        MIARN & 72.71 & 72.31 & 63.44 & 62.12 & 55.74 & 58.95 & 86.14 & 92.25\\
        Self-Atten & 76.01 & 75.19 & 66.29 & 65.47 & 61.32 & 60.12 & \underline{86.99} & \underline{93.48}\\
        \hline
        SIRM & \textbf{82.54$^*$} & \textbf{82.38$^*$} & \textbf{70.01$^*$} & \textbf{69.94$^*$} & \textbf{63.01$^*$} & \textbf{62.13$^*$} & \textbf{88.18$^*$} & \textbf{93.94$^*$}\\
        \hline
    \end{tabular}
\end{table}

NBOW realizes a decent performance for all datasets, especially for Reddit/movies. More importantly, NBOW has the lowest parameter size and achieves the least time cost. That means the NBOW can be a good choice in the vast majority of cases, also demonstrated by~\cite{shen2018baseline,conneau2018you}.

Unfortunately, the standard text representation models, such as CNN, LSTM, Atten-LSTM, and GRNN, don't outperform NBOW significantly. And they can not even achieve a stable performance across all datasets because of the lack of the training data. For example, the GRNN performs well on Tweets/gosh and Industry/spam, but works worse on Reddit/movies and IAC/v1. The state-of-the-art models SIARN, MIARN, and Self-Atten don't perform well as expected in this work intuitively. With more parameters and running time cost, these models may be even worse than NBOW. Moreover, RNN based models take more time than other models. 

The proposed SIRM significantly outperforms all the baseline models according to accuracy and F1 score. It is clear that the proposed SIRM, along with SRC, IRC, and Adv, can be more stable on all datasets which have the diverse data sizes and text lengths, with architecture specially designed to simulate human's reading comprehension process. Furthermore, other recent advanced models do not perform well due to the indifference of contextual information (for the word/sentence) and the bad impact of figurative expression. For example, SIRM makes it to identify \textit{ [Tweets/ghosh: sarcasm] `Do you know what I \underline{love}? Apartment construction at \underline{7 a.m}. \underline{3 mornings in a row}!'}, but SIARN and Self-Atten fail to do so. The reason may be that words in the last two sentences look irrelevant or not explicitly contradictory to the first sentence, which will mislead the two models. But, the SIRM can capture the real meaning by reading each word/sentence with the global knowledge.

It's worth mentioning that the parameter size and running time cost of the SIRM is comparable with all baselines. That is because we haven't used any recurrent unit which means the SIRM can be totally parallel during training and testing.

\begin{figure}[!ht]
\centering
\includegraphics[width=10cm]{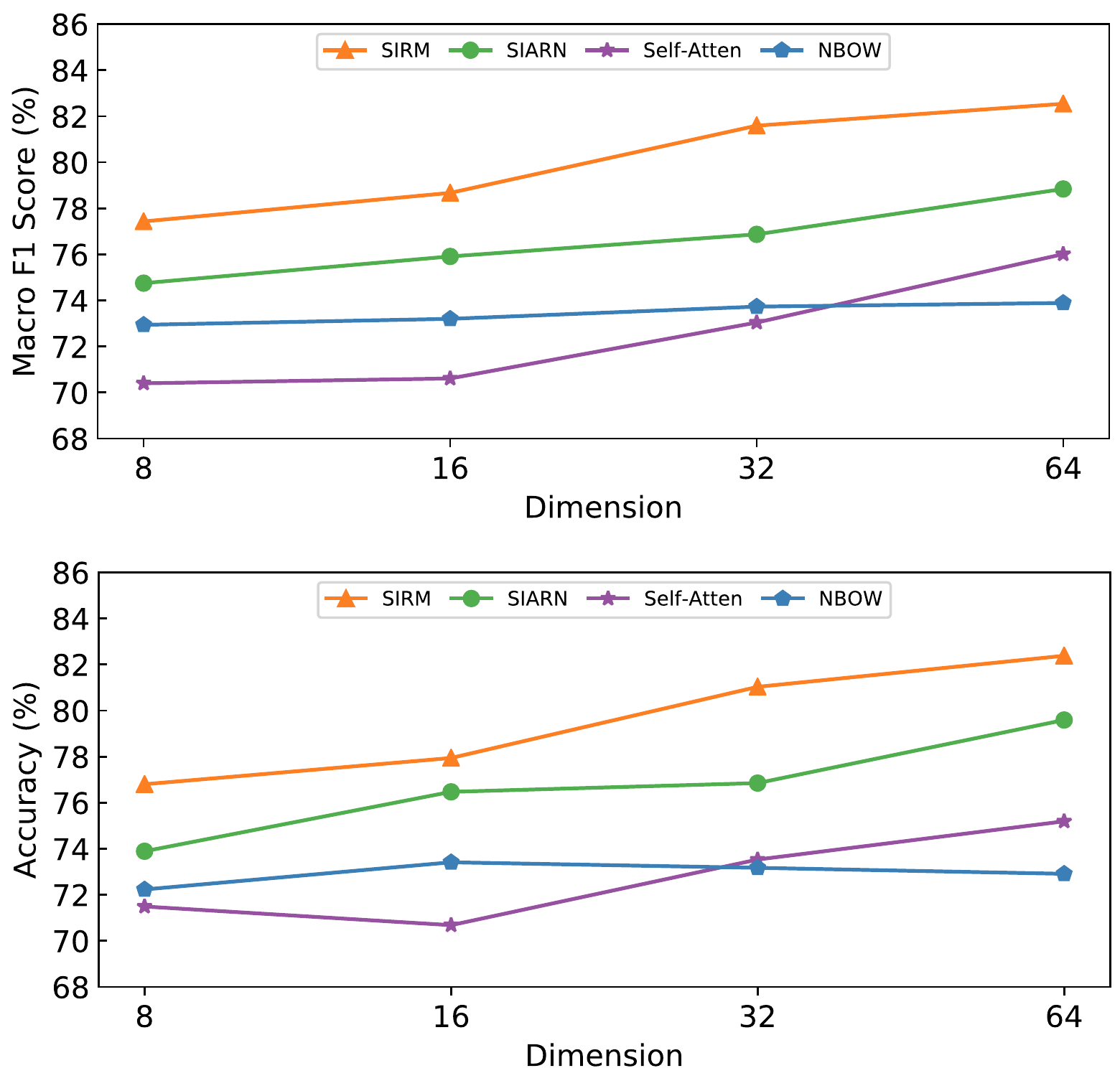} 
\caption{Results of comparison for parameter size sensitivity: x-axis is main dimension of the model.}
\label{fig:param}
\end{figure}

\subsection{Parameter Size Sensitivity}

As shown in Figure~\ref{fig:param}, parameter size sensitivity of the proposed SIRM against other baseline models is investigated based on Tweets/gosh. It is obvious that the proposed SIRM outperforms all representative baseline models, especially the tailored and state-of-the-art model for sarcasm detection, SIARN, according to macro F1 score and accuracy with all alternative parameter sizes. In contrast, Self-Atten achieves a lower score even than NBOW at the lowest dimension setting while the NBOW reaches the most stable performance among all alternative dimension settings. All these evidences demonstrate the robustness and superiority of the proposed SIRM. 

\begin{figure}[!ht]
\centering
\includegraphics[width=10cm]{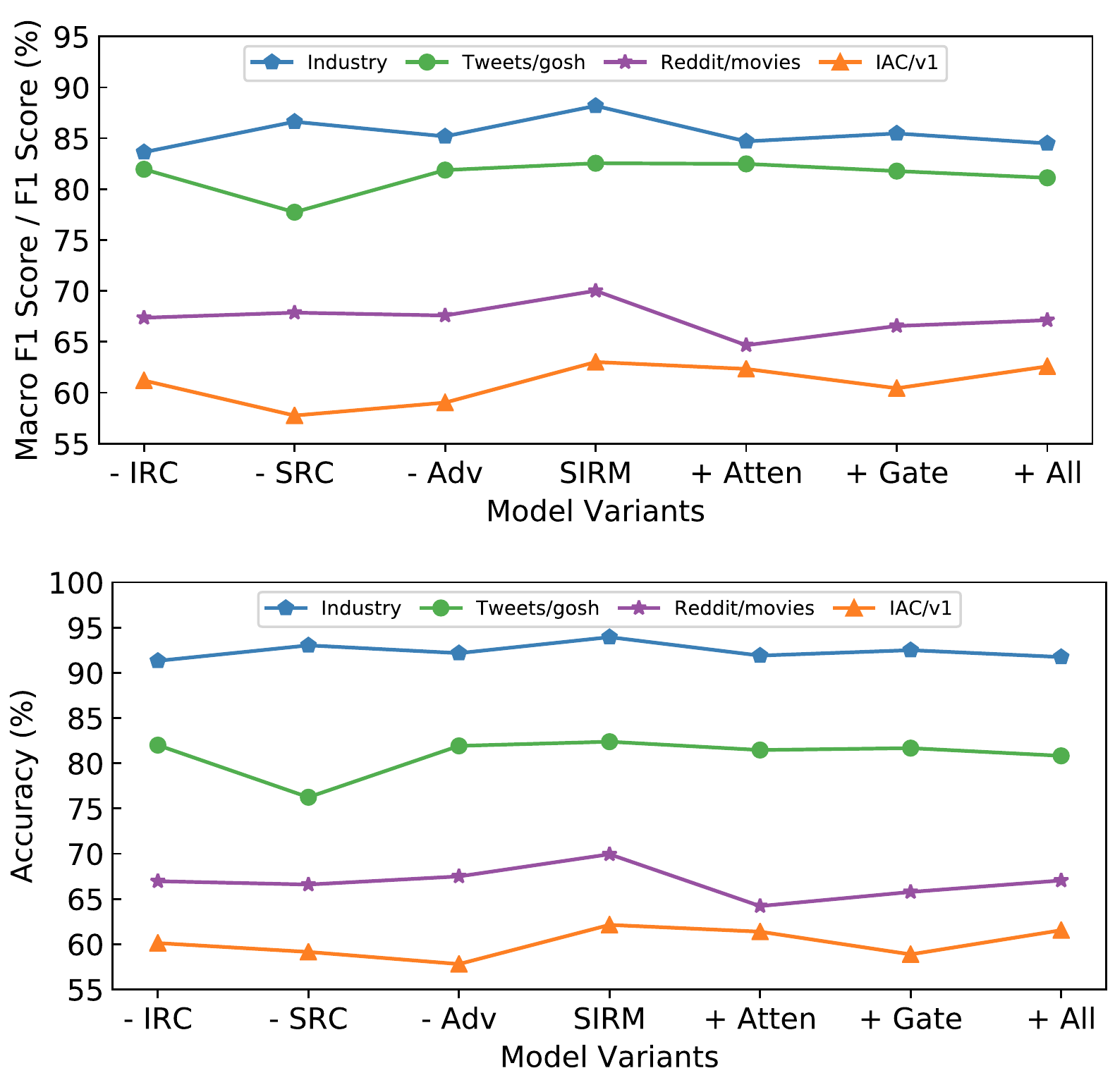} 
\caption{The performance of ablation and addition for the SIRM: - denotes the ablation and + denotes the addition.}
\label{fig:sub_and_add}
\end{figure}

\subsection{Ablation and Addition of SIRM}

For efficiency purpose, we design each part of the SIRM with respect to Occam's razor. Hence, we investigate the impact for complexity of the SIRM shown in Figure~\ref{fig:sub_and_add}. By removing each part, such as IRC, SRC, and Adv, there is a decrease compared with the SIRM. That is because each part of the SIRM plays a different, necessary, and important role for implied semantic meaning understanding across different datasets. Meanwhile, we find that using a more sophisticated component to replace the simple one is not advisable. Gate and attention mechanisms don't make performance increase. In particular, the more complex the model, the more space it will take. 

%% file: ipm-06-conclu.tex
\section{Conclusion}
\label{sec:conclu}
In this study, we propose a novel model, SIRM, for understanding and identifying the implied textual meaning with a quick manner. In SIRM, the SRC is designed to capture the dynamic global information, while the IRC is employed to characterize the fine semantics via a hierarchical framework by taking the contextual information into consideration with the dense connection. In addition, the adversarial loss is applied over the SRC to eliminate the potential noise. We conduct extensive experiments on several sarcasm benchmarks and an industrial spam dataset with metaphor. The results indicate that the proposed model practically outperforms all alternatives, in the light of performance, robustness, and efficiency.

%% file: ipm-08-acknowledge.tex
\section{Acknowledgments}
\label{sec:ack}

This work is supported by National Natural Science Foundation of China (71473183, 61876003) and Fundamental Research Funds for the Central Universities (18lgpy62).